\documentclass[runningheads]{llncs}
\usepackage[nameinlink,capitalize]{cleveref}
\usepackage{graphicx}
\usepackage{tabularx,array,booktabs}
\newcolumntype{L}{>{\centering\arraybackslash}m{3cm}}
\begin{document}
\title{Social Behaviour Understanding using Deep Neural Networks: Development of Social Intelligence Systems}
\titlerunning{Social Behaviour Understanding using Deep Neural Networks}
%
\author{Ethan Lim Ding Feng\inst{1} \and
Zhi-Wei Neo\inst{1} \and
Aaron William De Silva\inst{1} \and
Kellie Sim\inst{2} \and
Hong-Ray Tan\inst{2} \and
Thi-Thanh Nguyen\inst{3} \and
Karen Wei Ling Koh \inst{4} \and
Wenru Wang \inst{5} \and
Hoang D. Nguyen\inst{1}}
\authorrunning{Lim et al.}
%
\institute{University of Glasgow, Singapore \\
\email{2427232L@student.gla.ac.uk, 2355362N@student.gla.ac.uk, 2355348D@student.gla.ac.uk, Harry.Nguyen@glasgow.ac.uk}
\and
Ngee Ann Polytechnic, Singapore \\
\email{\{s10163148,s10177638\}@connect.np.edu.sg}
\and
National College for Education, Vietnam \\
\email{thanhtw76@gmail.com}
\and
National University Health System, Singapore \\
\email{karen\_wl\_koh@nuhs.edu.sg}
\and
National University of Singapore, Singapore \\
\email{nurww@nus.edu.sg}}
\maketitle              
\begin{abstract}

With the rapid development in artificial intelligence, social computing has evolved beyond social informatics toward the birth of social intelligence systems. This paper, therefore, takes initiatives to propose a social behaviour understanding framework with the use of deep neural networks for social and behavioural analysis. The integration of information fusion, person and object detection, social signal understanding, behaviour understanding, and context understanding plays a harmonious role to elicit social behaviours. Three systems, including depression detection, activity recognition and cognitive impairment screening, are developed to evidently demonstrate the importance of social intelligence. The study considerably contributes to the cumulative development of social computing and health informatics. It also provides a number of implications for academic bodies, healthcare practitioners, and developers of socially intelligent agents.

\end{abstract}
\newpage
\section{Introduction}
The landscape of social computing has evolved, with the proliferation of smaller, more powerful devices, such as mobile phones, tablets and wearable devices. Having all these advanced technologies available at our fingertips, people are now using them in everyday life, introducing new habits and generating new forms of data. 

The social computing paradigm has been moving beyond capturing information toward focusing on social intelligence \cite{wang2007social}. As a vital facet of human intelligence, social intelligence is the capability to understand oneself and to understand others. The development of social intelligence systems entails recognising social and behavioural patterns from the new types of data and providing in-depth analysis of social signals for better human support. This paper aims to address major boundaries of social computing capabilities and social signal processing by introducing a social behaviour understanding platform with the use of deep neural networks.

With the rapid advancement of artificial intelligence (AI), deep learning utilises complex networks of artificial neurons to provide new ways to investigate human interactions in various contexts. We propose a deep learning framework for understanding social signals and behaviours from an individual or a group of people.

The niche nature of the previous generations of approaches and devices restricted the types and possibilities for social behaviour analysis. With state-of-the-art technologies today, we introduce the design and implementation of social intelligence systems for activity recognition, behavioural analysis, and health assessment. The paper demonstrate three use cases of social intelligence.

\begin{itemize}
\item \textbf{Depression detection} aims to develop a social intelligence system, that uses machine learning techniques, to classify vocal features present in a depressed individual's voice. Utilising smartphone microphones, to determine if an individual suffers from depression through a mobile application.
\item \textbf{Activity recognition} aims to utilise smartphones, activity trackers and smartwatches, to collect accelerometer sensor data, for the classification of human activities. Proceeded by machine learning and deep learning techniques, to predict patient activities, for a fall prevention mobile application.
\item \textbf{Cognitive impairment screening} aims to build a tool to assess cognitive disorders based on individual's writings and movements with the use of convolutional neural networks (CNN). 
\end{itemize}

Based on academic foundations, the study contributes to the cumulative development of social intelligence and mobile health. It draws out many implications for academic theorists and healthcare practitioners.

The structure of the paper is as follows. Firstly, we review the literature background of our study in Sect. 2. Next, we present our social behaviour understanding architecture with the design concepts and three use cases of social intelligence systems. Lastly, the paper is concluded with findings and contributions.

\section{Literature Background}

\subsection{Social Intelligence}
Social intelligence is the ability to detect, interpret and react to human social and behavioural cues.
These cues have many facets, ranging from physical appearance to vocal and facial features and gestures.
Innately present in humans, social intelligence is a vital skill for understanding human behaviour, attributing significant impact on people's lives, to this form of intelligence \cite{albrecht2006social}.

With the advance of Internet technologies, social computing has been moving beyond social informatics towards emphasising on social intelligence \cite{wang2007social}. The field of implementing social intelligence in computers, is named social signal processing (SSP) \cite{vinciarelli2009social}. Computers are relatively untrained to comprehend the aforementioned social signals in most current applications. With existing computing capabilities, context-independent tasks like arithmetic and retrieval operations can be performed without issue; however, the current state of computing is struggled to handle context-dependent tasks, such as virtual-reality applications. Incapable of realising the full potential of Internet-of-Things (IoT) networks as it is unable to utilise the data generated to predict actions or needs \cite{vinciarelli2009social}.

\begin{figure}[]
\includegraphics[width=\textwidth]{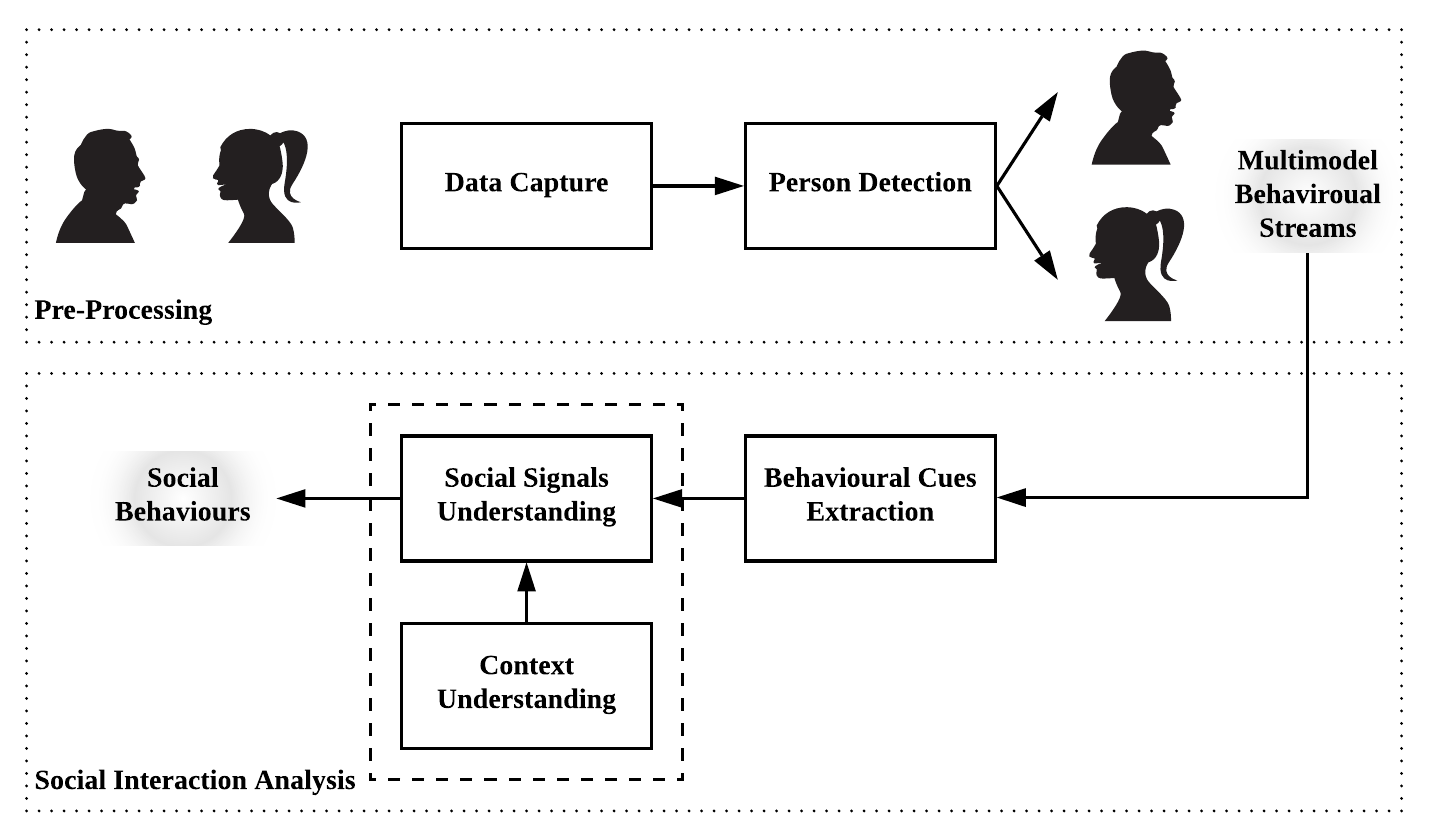}
\caption{Machine Analysis of Social Signals (Vinciarelli et al., 2009) \cite{vinciarelli2009social}} \label{fig:ssp}
\end{figure}

On the other hand, in regards to the ability to observe social signals, human tend to have fluctuating performance, whereas computers have more consistent performance. This indicates that humans are not fully utilising present social and behavioural cues, relying on more scenario-oriented contextual cues. While machines are able to utilise the cues more extensively \cite{pollick2002estimating,gold2008efficiency}. Vinciarelli et al. (2009) proposed a popular framework for machine analysis of social signals and behaviours as shown in \textbf{Fig. \ref{fig:ssp}.}

Achieving social intelligence will open up opportunities to a whole myriad of new applications. The development of social intelligence systems, therefore, becomes essential to stimulate greater availability of approaches and methodologies. To enable researchers and administrator to select the optimal approach, a guideline with clear objective and procedures will be invaluable.

\subsection{Machine Learning for Cognitive and Social Behavioural Detection}
The ever-growing popularity of artificial intelligence has led it to be applied in numerous fields of study. Empowering the discovery of novel applications, and thoroughly testing its limits. This trend has drawn focus into the utility of machine learning in health care \cite{nguyen2016automated}.

Many researchers have investigated the accuracy and viability of incorporating or utilising machine learning, with existing methodologies. The research results have proven capability of artificial intelligence at diagnosing various ailments and disorders, displaying high levels of accuracy, with the opportunity for further improvement \cite{rutkowski2019cognitive}. Wall et al. (2012) demonstrated an opportunity to improve healthcare methodology for diagnosing mental disorders and decrease healthcare costs \cite{wall2012use}. The use of AI in computer games has also been explored to evaluate human behaviour \cite{hildmann2018designing}. It was bound to specific behavioural preferences with the ability to simulate a certain degree of human behaviours. 

\begin{figure}[]
\includegraphics[width=\textwidth]{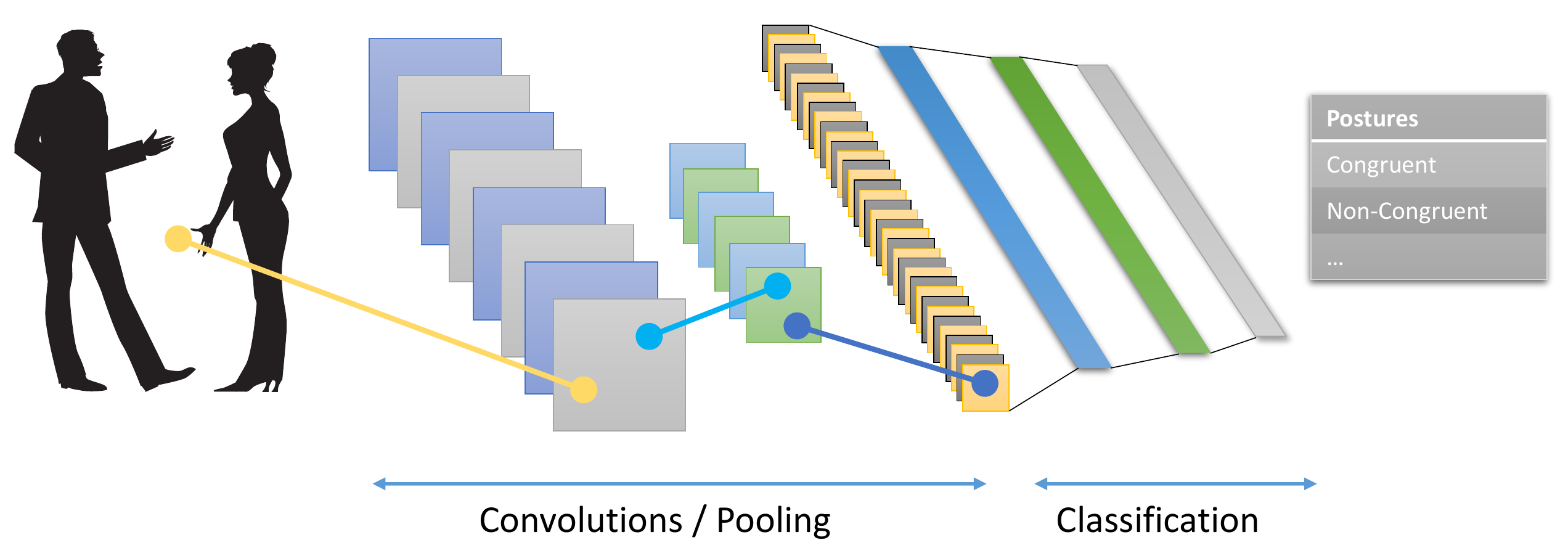}
\caption{Convolutional Neural Networks for Posture Detection} \label{fig:cnn}
\end{figure}

With recent breakthroughs in AI, deep learning has been well recognised as the next suitable wave of machine learning for social and behavioural analysis. Deep learning a multi-layered neural network, to steadily draws higher-level features from the input. Each deep neural network layer transforming the data to increasingly abstract representations of the input. A common deep learning implementation of interest is convolution neural networks (CNN), often used for image and video analysis \cite{lecun2015deep}, as shown in \textbf{Fig. \ref{fig:cnn}}. The CNN architecture consists of thousands to millions of artificial neurons in multiple layers, including convolutional layers, pooling layers, or fully connected activation layers. CNNs have been proven to perform with the better efficiency when it comes to vision and speech classification tasks, as shown in \cite{krizhevsky2012imagenet,hershey2017cnn}

\section{Social Behaviour Understanding using Deep Neural Networks}

\begin{figure}[]
\includegraphics[width=\textwidth]{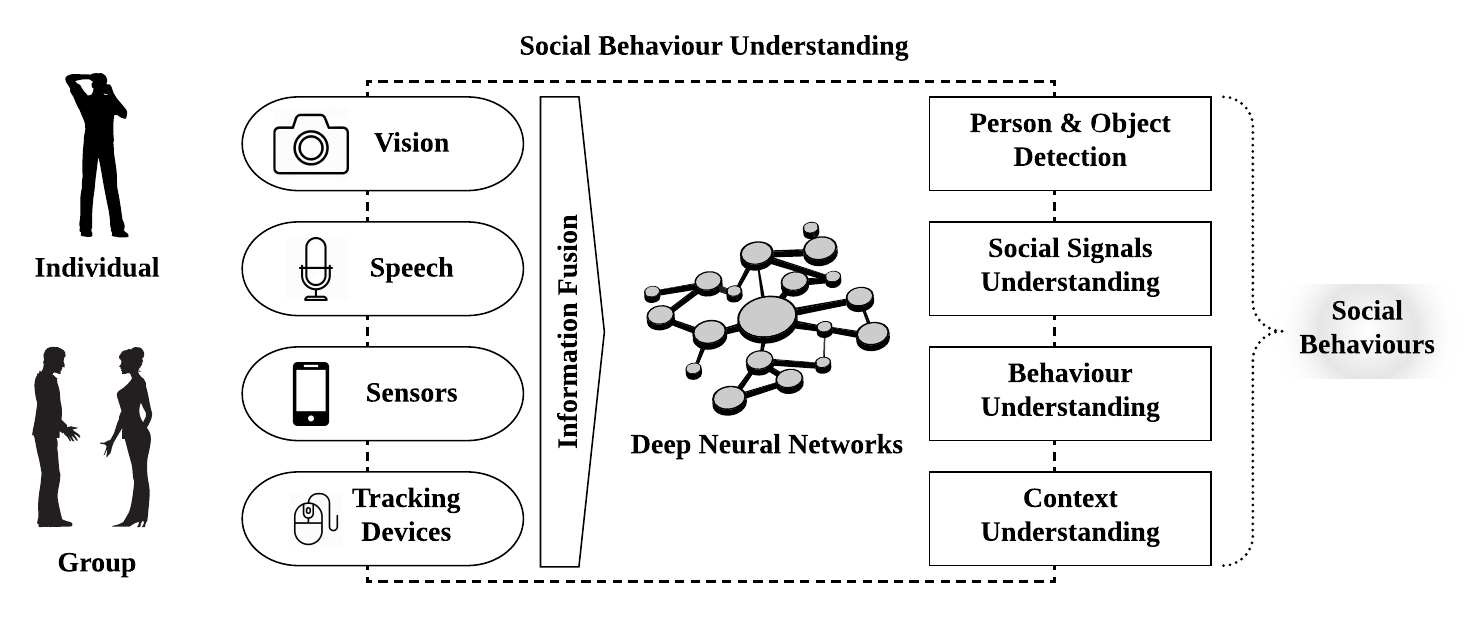}
\caption{Social Behaviour Understanding using Deep Neural Networks} \label{fig:sbu}
\end{figure}

In recent research, the use of artificial intelligence has been widely exploited to analyse the behavioural and social cues in social interactions \cite{wall2012use,wall2012use,hildmann2018designing}. This research trend has led to better understandings of human beings, where new knowledge and patterns of behavioural, social, and contextual cues are constantly discovered. With the new computing capabilities, we propose a framework for social behaviour understanding using state-of-the-art deep learning, as shown in \textbf{Fig. \ref{fig:sbu}}. Multiple constructs are adapted based on the original framework for machine analysis of social and behavioural signal processing from Vinciarelli, Pantic and Bourlard (2009) \cite{vinciarelli2009social}.

In our framework, the role of socially intelligent agents has been evolved to more closely emulate humans, thereby shortening the gap between machines and humans. We emphasise on fully realising the capabilities of artificial intelligence for robust and versatile detection of social and behavioural signals. The use of deep neural networks aims to simulate cerebral activities to create new ways of understanding data and making inferences. The proposed framework consists of FIVE (5) key components: (i) Information Fusion, (ii) Person and Object Detection, (iii) Social Signal Understanding, (iv) Behavioural Understanding, and (v) Context Understanding.

\begin{itemize}
\item \textbf{Information Fusion}. With the increasing ubiquity of Internet-of-Things (IoT) technology, new types of sensors, tracking devices, and mobile equipment have been widely introduced \cite{nguyen2018gamification}. These capabilities allow data capture of multimodal inputs, including visual, audible and movement data. Information fusion strategies are required to eliminate uncertainty and reliability issues in such data. The process of information fusion integrates multiple data sources into a robust, accurate and consistent input body for deep learning. Lee et al. (2008) suggested a hierarchical decomposing method to handle the data at three different levels: raw sensor data fusion, feature level fusion, and decision level fusion \cite{lee2008issues}. \\

\item \textbf{Person and Object Detection}. Social intelligence entails interactions among multiple agents, including humans and objects. The traditional methods in person and object detection are typically developed based on limited feature extraction and shallow learning models \cite{zhao2019object}. The recent breakthroughs in deep learning have raised a new ground for detecting objects with high confidence in audios, images and videos. Convolutional neural network models perform distinguishably, with a variety of network architectures, training and optimisation strategies. It is also important to note that detection models can be integrated within a single multimodal neural network architecture. \\

\item \textbf{Social Signal Understanding}. Social signals occur in everyday situations, which include many social cues such as attention, empathy, politeness, or agreement. Social signal processing has drawn huge research efforts to understand human interactions in an automated and continuous manner \cite{schuller2013interspeech,likamwa2013moodscope,gunes2010automatic}. Deep evolutional spatial-temporal networks were suggested to extract both temporal and spatial features of facial expressions, which outperformed traditional approaches in a large margin \cite{zhang2017facial}. Similarly, deep learning has been used to learn social signals from appearance, gesture and posture \cite{chen2019analyze}. \\

\item \textbf{Behavioural Understanding}. Human behaviours play a vital role in shaping the perception of human interactions. Investigating behavioural cues, hence, allows intelligent agents to elicit social signals with a higher degree of support. This is also applicable to individual behaviours, captured with or without social interactions, due to temporal dynamics of social behaviours. This framework suggests behavioural understanding component is a good supplement to develop social intelligence. \\

\item \textbf{Context Understanding}. Understanding social and behavioural signals is not without contextual information such as location, time, or situation. The contexts are tightly associated with communicative intention; thus, it is critical to consider their dynamics in social behaviour analysis. With new mobile and sensor capabilities, the presence of context data can be embedded into multimodal deep neural networks in various ways \cite{vahora2018group}. \\ 
\end{itemize}

With the recent development in deep neural networks, the fusion of multimodal understanding units opens new pathways to analyse and recognise social behaviours. Frequently, social, behavioural, and contextual dimensions of the data contain both unique and overlapped signals; thus, training using deep neural networks is a viable option for the development of intelligent agents. Cross-modality transformers are increasingly explored to address the challenge of multifacet representation learning and pattern recognition \cite{tan2019lxmert}, such as social intelligence.

\section{Development of Social Intelligence Systems}

Based on the Social Behaviour Understanding framework, the study takes an important step to develop three social intelligence systems for health assessment. They utilise deep neural networks to detect social and behavioural cues using real-time data for timely interventions.

The paper aims to bring collaborative care to the next level, where social behaviours are recognised and exchanged with social support agents. 

\subsection{Use Case 1: Depression Detection}

Current methods of diagnosing depression are time-consuming and archaic, with only minor improvements being made regarding its process \cite{beck1961inventory}. The process requires psychiatrists to initially screen patients through questionnaires utilising scales, including but not limited to: Centre for Epidemiological Studies Depression Scale (CES-D) \cite{spitzer1999validation}, Beck Depression Inventory (BDI) \cite{lewinsohn1997center} or PRIME-MD Patient Health Questionnaire \cite{kroenke2001phq,kroenke2002phq}. 

\begin{figure}
\centering
\includegraphics[scale=0.4]{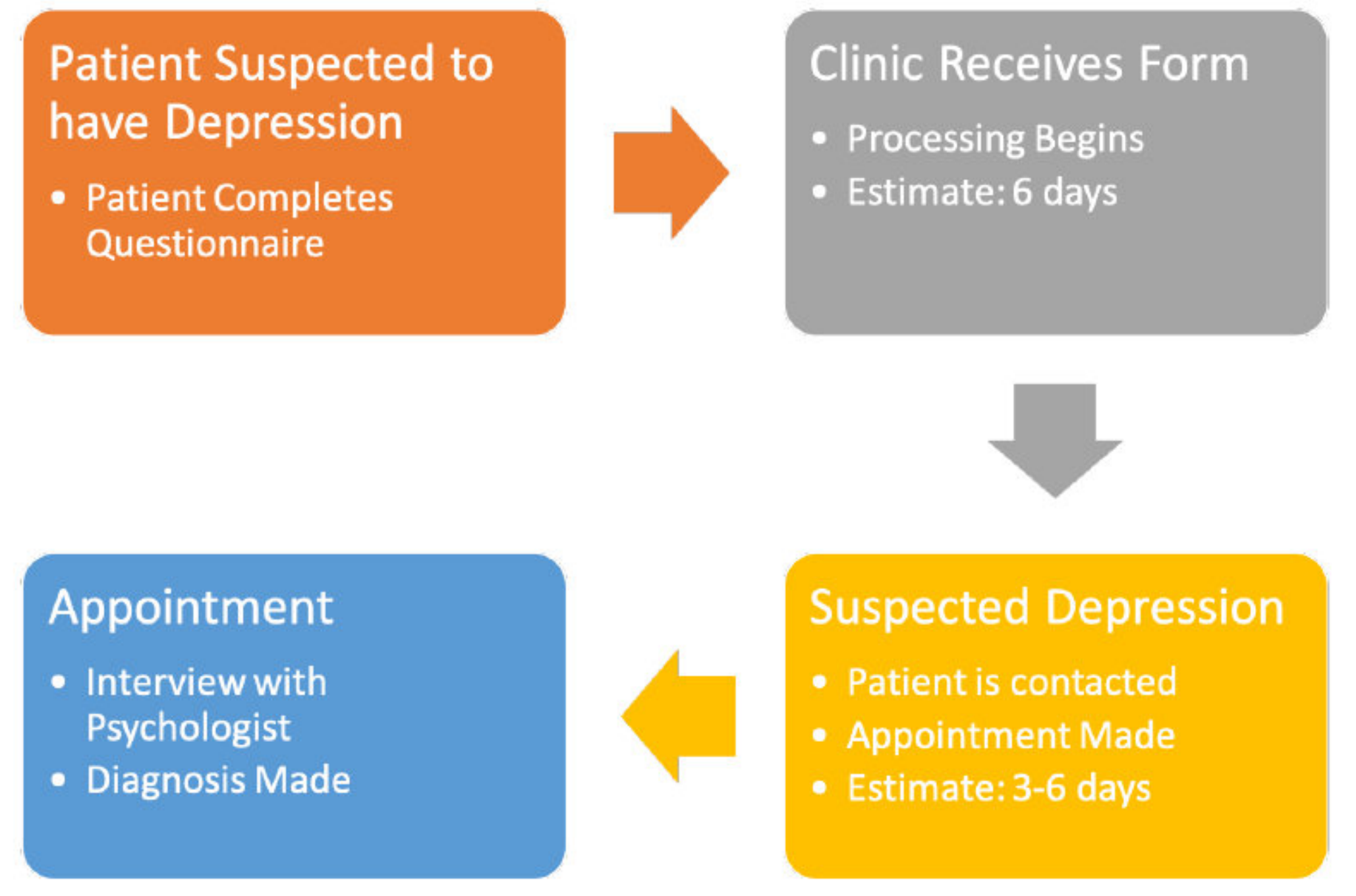}
\caption{Process of Diagnosing Depression. \cite{beck1961inventory}} 
\label{fig:depressionprocess}
\end{figure}

After screening, individuals suspected of suffering from depression would be contacted to arrange for an additional appointment to affirm the diagnosis \cite{beck1961inventory}.  This process could entail a minimum wait period of 2 weeks before patients are diagnosed and can begin treatment, as depicted in \textbf{Fig. \ref{fig:depressionprocess}}. Currently, much of the waiting period, is devoted to the processing of questionnaires, and the scheduling for an appointment. Each of the patient’s responses must be evaluated by a psychologist, after which, the result is only an indication of whether the patient suffers from depression.

Few studies and experiments have been conducted to evaluate the effectiveness of speech-based depression detection. However, 2 studies, Depression Speaks \cite{eyben2016open} and Depression Detect \cite{scibelli2018depression} used, The Distress Analysis Corpus -Wizard of Oz (DAIC-WOZ) database \cite{gratch2014distress}, to experiment with speech-based depression detection.  Utilising machine and deep learning, respectively, to extract features and classify depression from speech audio. Nevertheless, there is no mobile application currently, that is able to detect symptoms of depression, based on their voice. 
By attempting to improve the medical industry through novel means, advances in the methods of mental health diagnosis could be made in the future.

This system aims to explore the feasibility of using a mobile application to detect patients with depression based on their vocal features. Allowing it to predict in real time, if an individual displays symptoms of depression.

\begin{figure}[]
\centering
\includegraphics[scale=0.6]{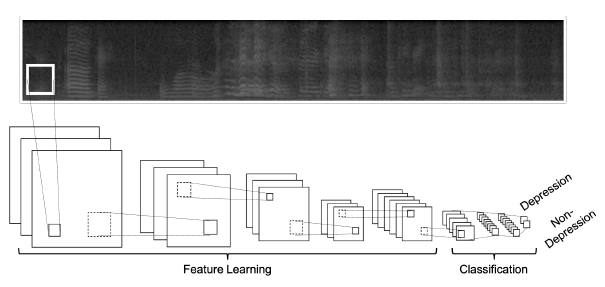}
\caption{Convolutional Neural Networks for Spectogram} 
\label{fig:spectrogram}
\end{figure}

\subsubsection{Data Collection.}
Data is collected from the built-in microphone of, in this case, a Google Pixel 2 XL. The output format is a Pulse Code Modulation (PCM) file, which is a file format that represents a digitization of analog audio. The sampling rate is 44100Hz which means that there are 44100 samples of audio frequency per second. 

The training dataset used for this project is The Distress Analysis Interview Corpus - Wizard of Oz (DAIC-WOZ) database by the University of Southern California (USC) \cite{gratch2014distress}.  Its contains 189 clinical interview sessions designed to support the diagnosis of psychological distress conditions.  Each session comprises of a transcript of the interview, an audio recording and the facial features of the participant.

\subsubsection{Development.}

\begin{figure}[]
\centering
\includegraphics[width=8cm]{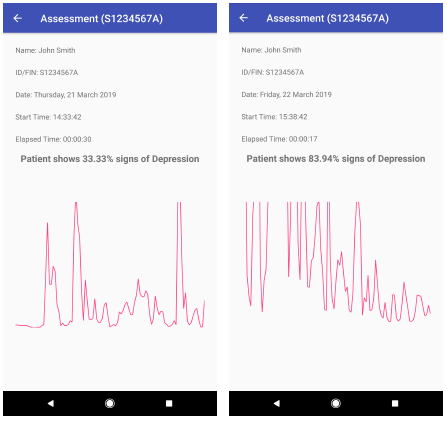}
\caption{Screenshots of the application processing live data of 2 different users.} 
\label{fig:depressprototype}
\end{figure}

We processed the data into visual representations using convolutional neural networks in \textbf{Fig. \ref{fig:spectrogram}}. And a prototype mobile application was developed using the Android Platform. Upon opening the application, the user is prompted to enter the Name and ID of the patient before beginning the session. Once a session is started, the user can start the recording process to detect possibility of having depression, as shown in \textbf{Fig. \ref{fig:depressprototype}}.

\subsection{Use Case 2: Activity Recognition for Fall Detection and Prevention}

\begin{figure}[]
\centering
\includegraphics[scale=0.4]{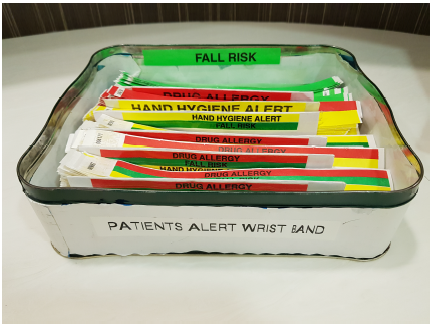}
\caption{Fall-risk wrist tag} 
\label{fig:riskfalltag}
\end{figure}

Patient accidents in hospitals are of significant concern, especially if they occur with the elderly. Globally, a third of adults over 65 years old, falls once a year. These accidents could lead to additional harm, such as further injury, complications and loss of mobility. Therefore, this social intelligence system aims to recognise patients' activities for fall detection and prevention.
High-risk patients are given green wrist tags, as shown in Fig. \ref{fig:riskfalltag}, and a green label, which are set at the panel of their beds, and they are required to always be continuously monitored in the system.

\subsubsection{Data Collection.}
The widespread proliferation of smartphones has made low-cost smartphones equipped with a variety of sensors commonplace. This project would explore the use of mobile technologies to enhance the fall detection and prevention strategy further. The training dataset from UniMiB-SHAR was used as it was an open dataset available online. The UniMib-SHAR dataset consists of 17 different kinds of activities, divided into nine different types of daily activities such as walking, running, etc. and eight different types of falls such as fall forward, fall left, etc. There are a total of 7759 daily activities, and 4192 falls respectively.

The social intelligence would be performed in real-time with the assistance of a smartphone, with an in-built accelerometer. The patient would carry a smartphone, with the mobile application deployed to it and perform different activities. Logged accelerometer data of 1-second intervals would be sent through an API call to the server for processing. 

\begin{figure}[]
\centering
\includegraphics[width=6.8cm]{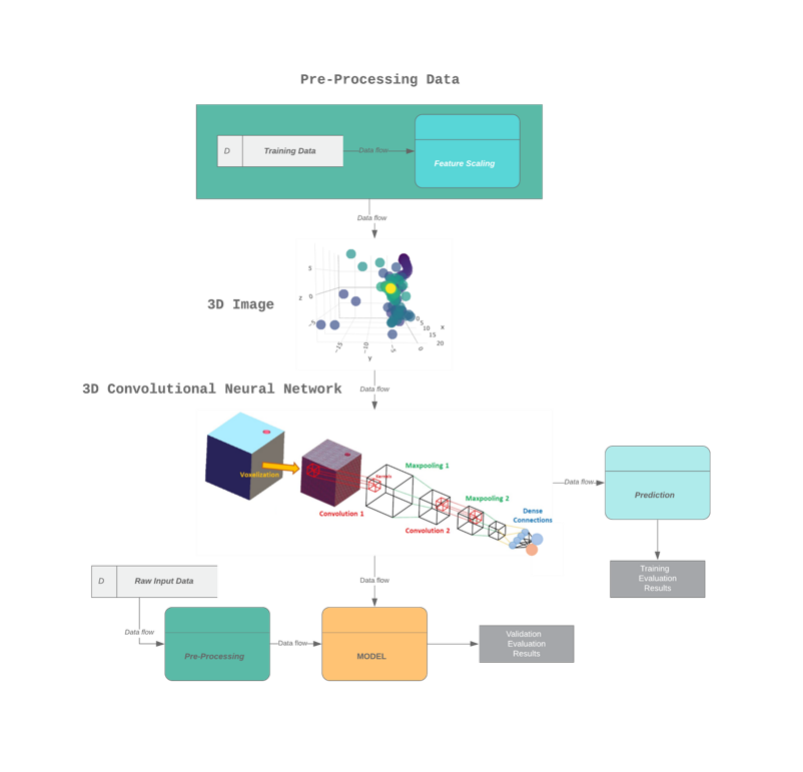}
\caption{3D Convolution Neural Networks for Behavioural Analysis} 
\label{fig:3dcnn}
\end{figure}

\subsubsection{Development.} \label{sssec:mdev}

We employed 3D convolutional neural networks to analyse the behavioural data, as shown in \textbf{Fig. \ref{fig:3dcnn}}. The 4D tensor would then be passed through the many convolutional, pooling, batch normalisation, flatten, and multi-perceptron layers to finally the activation layer. This would generate the 3D CNN model and show the training accuracy of the model. Firebase was used to provide real-time database as a backend service to store and return the information of the patient's name, activity and time of activity to be displayed on the clinician and patient applications.

\subsection{Use Case 3: Cognitive Impairment Screening}
75 million people are predicted to be affected by dementia by 2030 \cite{world2017global}. With individuals older than 65 years, at much greater risk of developing a form of cognitive impairment. Hospitals employ a battery of cognitive tests, to detect cognitive impairments. The tests commonly take the form of writing and drawing examinations, requiring the completion of tasks ranging from simple instructional writing, to complex memory-based drawings.

This social intelligence system aims to predict the risk of cognitive impairments with the use of hand writings and pen movements. 

\begin{figure}[]
\centering
\includegraphics[scale=0.20]{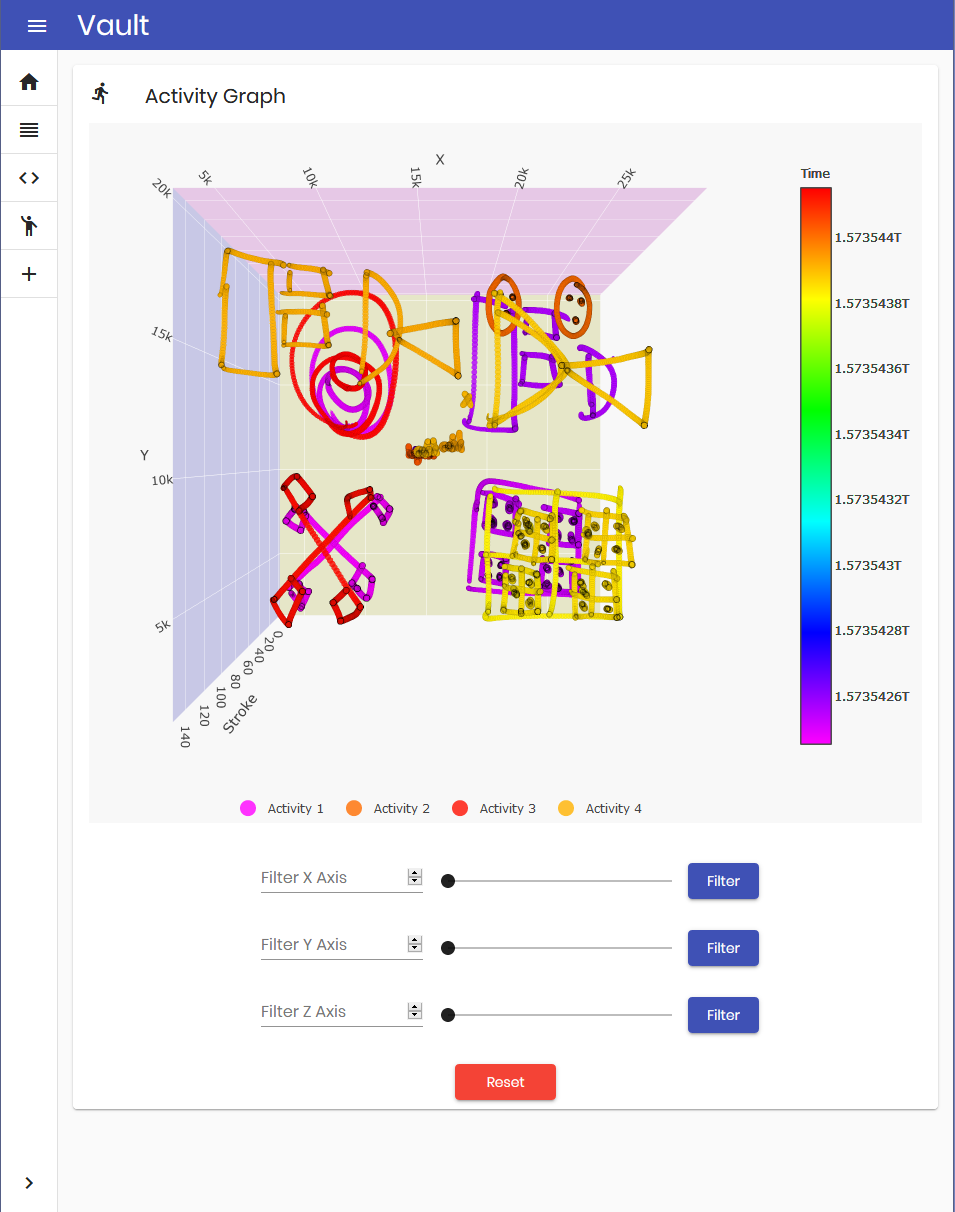}
\caption{3D Visualisation of Subject Data} 
\label{fig:demprototype2}
\end{figure}

\subsubsection{Data Collection.}
In this study, participants undergo a series of cognitive manual handwriting tests; and electronic tablet and pen was used to capture writing and in-air (hover) trajectory.

\subsubsection{Development.}
We developed 3D images from feature scaling the training and test data, will be represented by a 4D tensor. Then, the tensors would then be passed through a 3D deep neural networks for cognitive impairment detection. In our development, Angular 8 and JavaScript framework was utilised for the frontend, Flask and Python for the backend, and MongoDB for the database. Finally, after configuration, a 3D model will created from the variables selected, with the option to filter the X, Y and Z axis, for inspection as shown in \textbf{Fig. \ref{fig:demprototype2}}.

\section{Conclusion}
Social intelligence systems are promising to revolutionise people's everyday life. Our study proposes a social behaviour understanding framework which performs recognition of social and behavioural signals for the development of socially intelligent systems. The framework consists of five key components: (i) Information Fusion, (ii) Person and Object Detection, (iii) Social Signal Understanding, (iv) Behavioural Understanding, and (v) Context Understanding. Cross-modality analysis of social, behavioural, and contextual information with the use of deep neural networks is suggested to bring social intelligence to the next level. Moreover, we developed three social intelligence systems for depression detection, activity recognition, and cognitive impairment screening. 

Our study contributes to the cumulative theoretical development of social computing and artificial intelligence. The uniqueness of social intelligence is evidently demonstrated to shed light on new applications. We hope our social behaviour understanding framework provides meaningful guidelines on the development of new types of social computing systems. This paper is not an end, but rather a beginning of future research as we are looking into ways of further refining and evaluating our social intelligence systems.



\bibliographystyle{splncs04}
\bibliography{reference}

\end{document}